\documentclass{article}

\usepackage{microtype}
\usepackage{graphicx}
\usepackage{subcaption}
\usepackage{booktabs} 

\usepackage{hyperref}

\newcommand{\bs}[1]{\mathbf{#1}}

\usepackage[preprint]{icml2026}

\usepackage{amsmath}
\usepackage{amssymb}
\usepackage{mathtools}
\usepackage{amsthm}
\usepackage{enumitem}
\usepackage{titlesec}

\usepackage[capitalize,noabbrev]{cleveref}

\theoremstyle{plain}
\newtheorem{theorem}{Theorem}[section]

\newtheorem{lemma}[theorem]{Lemma}

\theoremstyle{definition}

\theoremstyle{remark}

\usepackage[textsize=tiny]{todonotes}

\icmltitlerunning{Information-Regularized Constrained Inversion for Stable Avatar Editing from Sparse Supervision}

\begin{document}

\twocolumn[
  \icmltitle{Information-Regularized Constrained Inversion for \\Stable Avatar Editing from Sparse Supervision}



  \icmlsetsymbol{equal}{*}

  \begin{icmlauthorlist}
    \icmlauthor{Zhenxiao Liang}{ut}
    \icmlauthor{Qixing Huang}{ut}
  \end{icmlauthorlist}

  \icmlaffiliation{ut}{Department of Computer Science, University of Texas at Austin}


  \icmlkeywords{Machine Learning, ICML}

  \vskip 0.3in
]



\printAffiliationsAndNotice{}  

\begin{abstract}
  Editing animatable human avatars typically relies on sparse supervision—often a few edited keyframes—yet naively fitting a reconstructed avatar to these edits frequently causes identity leakage and pose-dependent temporal flicker. We argue that these failures are best understood as an ill-conditioned inversion: the available edited constraints do not sufficiently determine the latent directions responsible for the intended edit. We propose a conditioning-guided edited reconstruction framework that performs editing as a constrained inversion in a structured avatar latent space, restricting updates to a low-dimensional, part-specific edit subspace to prevent unintended identity changes. Crucially, we design the editing constraints during inversion by optimizing a conditioning objective derived from a local linearization of the full decoding-and-rendering pipeline, yielding an edit-subspace information matrix whose spectrum predicts stability and drives frame reweighting / keyframe activation. The resulting method operates on small subspace matrices and can be implemented efficiently (e.g., via Hessian-vector products), and improves stability under limited edited supervision.
\end{abstract}

\begin{figure}[t!]
  \centering
  \includegraphics[width=0.9\linewidth]{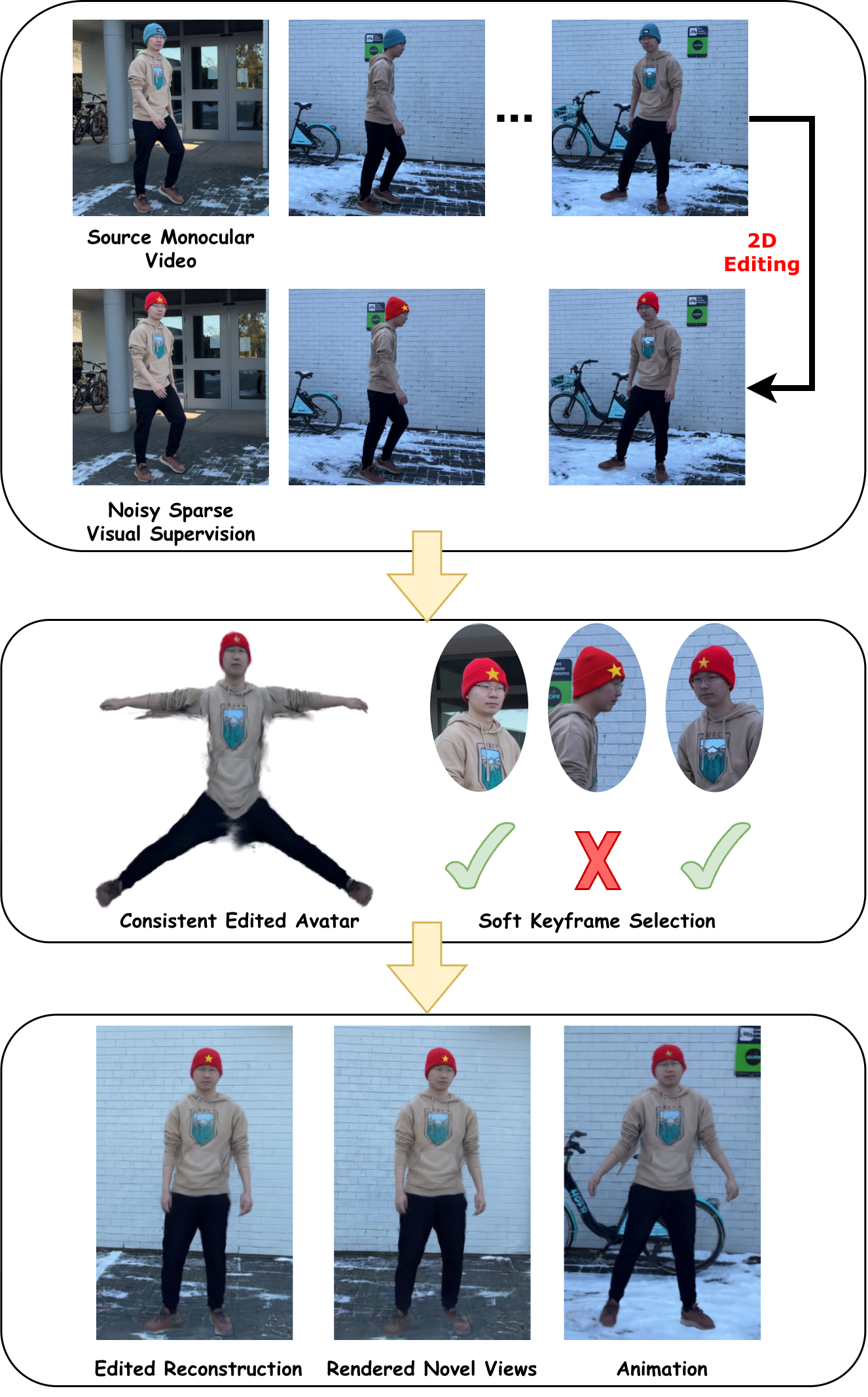}
  \caption{\textbf{Task Overview.} Given a source monocular video (1st row) and sparse edited keyframes (2nd row), our method produces temporally stable avatar edits that preserve identity, along with per-keyframe importance weights (3rd row). The edited avatar supports downstream applications such as novel view synthesis and animation (4th row).}
\vspace{-0.2in}
  \label{fig:teaser}
\end{figure}

\section{Introduction}
\label{sec:intro}

Reconstructing high-quality animatable 3D human avatars from sparse observations, such as a short monocular clip or a small set of images, is an increasingly popular task in visual computing. Recent progress is driven by explicit, differentiable representations---notably 3D Gaussian Splatting (3DGS)~\cite{kerbl2023gaussian}---combined with pose-driven deformation, enabling photo-realistic rendering and animation. A growing set of work reconstructs animatable human Gaussians from monocular or sparse-view videos by learning a canonical representation together with pose-conditioned deformation~\cite{qian20243dgsavatar,hu2024gauhuman,moreau2024hugs,li2024animatablegaussians,jiang2023instantavatar}.
Many pipelines further exploit structured canonical spaces anchored to a parametric body model (e.g., SMPL)~\cite{loper2015smpl} to improve pose generalization and local control, and some incorporate learned priors to make inversion and reconstruction robust under sparse and inconsistent observations~\cite{buhler2025dla,dong2025moga,wang2025tera}.

Although reconstruction has advanced rapidly, practical applications frequently call for a desired property: \emph{edited reconstruction}.
Here, a user specifies an appearance edit (e.g., garment or shoe replacement) on only a few keyframes, and the system must propagate the edit consistently to the entire pose sequence without distorting identity or introducing temporal flicker.
A common approach is to apply strong 2D diffusion-based editors to obtain edited images and then refit a 3D representation to match them, often with additional multi-view consistency constraints~\cite{wang2024gaussianeditor,palandra2024gsedit,chen2024dge,haque2023instructnerf2nerf,dong2023vicanerf}.
Interactive pipelines may treat key views as a user control handle to steer editing strength and preference~\cite{wen2025intergsedit}.
However, in the sparse-keyframe regime, naively fitting a reconstructed avatar to a small number of edited frames routinely exhibits two failure modes:
(i) \textbf{identity leakage}, where optimization drifts beyond the intended edit region and alters identity-defining attributes; and
(ii) \textbf{pose/time inconsistency}, where the edit looks plausible on edited keyframes but degrades under pose change, producing view-dependent artifacts and frame-to-frame flicker.

We argue that these failures are not merely implementation issues, but symptoms of an \emph{ill-conditioned inversion}.
Edited reconstruction typically starts from a base identity latent \(z_{\text{base}}\) obtained by inversion, and editing updates this latent to satisfy sparse edited constraints.
With only a few edited keyframes, many latent directions can explain the supervised frames equally well while behaving very differently under unseen poses, making the inverse problem sensitive to noise, mask imperfections, and modeling mismatch.
This motivates the desired goal in which
\emph{edited reconstruction should be treated as a constrained inversion problem whose stability is governed by inversion conditioning; crucially, the editing constraints themselves should be designed during inversion.}

\textbf{Inversion conditioning for edited reconstruction.}
We propose a framework that performs editing as a constrained inversion in a structured avatar latent space.
First, we restrict edits to a low-dimensional, part-specific \emph{edit subspace} (e.g., garments or shoes), which reduces degrees of freedom and limits leakage.
Second, rather than treating keyframes as a heuristic pre-processing choice, we associate each candidate frame with a continuous constraint weight and optimize these weights jointly with the edit code.
Our key tool is a conditioning objective derived from a local linearization of the full decoding-and-rendering pipeline, yielding an \emph{edit-subspace information matrix} whose spectrum predicts inversion stability.
By maximizing a spectral conditioning criterion (e.g., log-determinant) inside the inversion loop, the method reallocates supervision toward the most stabilizing constraints; in this view, ``keyframes'' \emph{emerge} as active constraints induced by inversion-time constraint design.
Although information-theoretic criteria are well-established for view selection and data acquisition in reconstruction~\cite{pan2022activenerf,jiang2023fisherrf,wilson2025popgs}, our use is different: we do not choose camera views to improve reconstruction coverage, but allocate \emph{sparse edited supervision} to stabilize an edit-code inversion system.

A practical advantage of our formulation is efficiency.
All conditioning quantities live in the low-dimensional edit subspace, so optimization operates on small \(r\times r\) matrices (with \(r\ll d\)) where $d$ is the latent dimension.
The required second-order regularization terms can be computed without constructing high-dimensional Jacobians, e.g., via Hessian-vector products, and cached per sequence.
Moreover, the edit subspace can be learned from a modest number of paired original/edited assets, enabling data-efficient part control without training a massive, universally general 3D editing model.

In summary, we make the following contributions:
\begin{itemize}[leftmargin=*,itemsep=0.2em]
\item \textbf{Edited reconstruction as stable constrained inversion.} We formulate sparse-keyframe avatar editing after reconstruction as a constrained inversion in a structured latent space rendered by a differentiable avatar pipeline.
\item \textbf{Inversion-time constraint design via conditioning.} We introduce an information-regularized optimization that jointly learns an edit code and per-frame constraint weights by optimizing the spectral conditioning of an edit-subspace information matrix, improving stability under sparse edited supervision.
\item \textbf{Efficient, data-light realization.} Our conditioning analysis operates on small subspace matrices and can be implemented efficiently (e.g., via Hessian-vector products and caching); the edit subspace can be learned from limited paired edits, reducing the need for heavy training resources.
\end{itemize}

\section{Related Work}
\label{sec:related}

\noindent\textbf{Animatable Human Avatars with Gaussian Splatting.} 3D Gaussian Splatting (3DGS) provides an explicit, differentiable representation that enables high-fidelity rendering at interactive rates~\cite{kerbl2023gaussian}.
Building on 3DGS, several methods reconstruct animatable humans from monocular or multi-view videos by learning canonical Gaussians together with pose-driven deformation fields~\cite{qian20243dgsavatar,hu2024gauhuman,moreau2024hugs,li2024animatablegaussians,jiang2023instantavatar}.
These methods primarily target \emph{reconstruction} (novel view/novel pose synthesis) and do not address the unique instability that arises when only a few frames receive explicit user edits.

Recent pipelines increasingly incorporate structured canonical spaces (e.g., UV-aligned maps anchored to a parametric body model such as SMPL) to improve pose generalization and enable local control~\cite{loper2015smpl,buhler2025dla}.
MoGA~\cite{dong2025moga} further highlights the value of a generative avatar prior and formulates monocular reconstruction as inversion into a learned latent space.
Similarly, TeRA~\cite{wang2025tera} studies text-guided avatar generation in a structured latent diffusion space.
In contrast, our goal is \emph{edited reconstruction}: given a base reconstruction/inversion, we propagate sparse keyframe edits across the pose sequence while preserving identity and suppressing flicker, without training a large, universally general 3D editing model.

\noindent\textbf{Garment-Aware and Layered Avatar Representations.}
A parallel line of work explicitly disentangles garments from the underlying body to support try-on and garment-level editing.
Layered Gaussian avatars separate clothing and body into different layers to improve tracking and facilitate clothing transfer~\cite{lin2024layga,gong2024laga}.
GGAvatar reconstructs garment-separated 3DGS avatars from monocular video through template-based separation~\cite{chen2024ggavatar}, while Disco4D models clothing assets as Gaussians on an SMPL-X body for disentangled 4D human generation and animation~\cite{pang2024disco4d}.
Gaussian-based virtual try-on further integrates 2D VTON signals with 3DGS editing pipelines to enable garment replacement~\cite{chen2024gaussianvton,cao2024gsvton}.
We do not claim novelty in garment separation or try-on \emph{per se}. Instead, our contribution is orthogonal: we study how to \emph{impose and allocate sparse edited supervision} so that a reconstruction-first avatar can be edited stably under new poses/time, even when the user only edits a few frames.

\noindent\textbf{Multi-View / Video-Consistent 3D Editing.}
A common paradigm for 3D editing is to obtain edited images using powerful 2D diffusion editors, then fit a 3D representation to match these edited views while maintaining cross-view consistency.
For 3DGS, GaussianEditor~\cite{wang2024gaussianeditor} and GSEdit~\cite{palandra2024gsedit} optimize Gaussians under diffusion guidance for text-driven edits, and EditSplat improves view-consistency via multi-view fusion guidance and attention-guided trimming~\cite{lee2025editsplat}.
DGE explicitly decomposes the pipeline into multi-view consistent 2D editing followed by direct 3DGS fitting for improved efficiency~\cite{chen2024dge}.
Analogous ideas exist for NeRF editing, where view-consistent constraints are propagated from key views using geometry-aware correspondences~\cite{haque2023instructnerf2nerf,dong2023vicanerf}.

Several very recent works emphasize sparse-view or single-frame guided 3D editing via foundation models.
Tinker produces consistent multi-view edits from one or two input images without per-scene optimization~\cite{zhao2025tinker}.
EditCast3D propagates a single-frame edit with video generation models and performs view selection before reconstruction~\cite{qu2025editcast3d}.
InstructMix2Mix distills a 2D diffusion editor into a pretrained multi-view diffusion model to improve consistency under sparse-view editing~\cite{gilo2025instructmix2mix}.
Bengtson et al.~\cite{bengtson2025diffusionguidance} enforce cross-view consistency during diffusion sampling via a training-free guidance loss.
SplatPainter focuses on interactive 3DGS authoring from 2D edits via feedforward updates and test-time training~\cite{zheng2025splatpainter}, while InterGSEdit introduces user-interactive key-view selection for controllable 3DGS editing~\cite{wen2025intergsedit}.

Our setting differs in both \emph{supervision} and \emph{mechanism}.
Rather than generating a fully edited multi-view dataset prior to reconstruction or distilling a multi-view diffusion model, we assume a reconstruction-first avatar (with a differentiable renderer) and treat editing as a \emph{constrained inversion} in a low-dimensional edit subspace.
Crucially, we optimize per-frame constraint weights \emph{within} inversion using a conditioning objective derived from a local linearization of the rendering pipeline.
This turns ``keyframes'' into active constraints that \emph{emerge} from inversion-time constraint design, directly targeting stability (reduced leakage and flicker) under sparse edited supervision.

\noindent\textbf{Information-Theoretic View Selection and Optimal Experimental Design.}
Information gain and optimal experimental design (OED) criteria (e.g., log-determinant or minimum eigenvalue objectives) are widely used for view selection in active reconstruction and uncertainty-aware radiance fields~\cite{pan2022activenerf,jiang2023fisherrf}.
POp-GS brings OED-style uncertainty quantification and next-best-view selection to 3DGS~\cite{wilson2025popgs}.
These works indicate that spectral criteria are well-established for \emph{data acquisition}.
Our use is different: we do not select camera views to improve reconstruction coverage.
Instead, we reuse the OED lens to analyze and improve the \emph{conditioning of an edit-code inversion system} in an avatar pipeline, and we allocate sparse edited supervision via differentiable weights to stabilize edited reconstruction.


\section{Method}
\label{sec:method}

We study \emph{stable edited reconstruction} of animatable human avatars. Given a base reconstructed avatar and sparse edited supervision (typically a few edited frames with localized masks), our goal is to produce a single edited avatar representation whose renderings match the desired edits across the entire sequence while avoiding identity leakage and pose-dependent temporal flicker.

Our key idea is to treat editing as a constrained inversion in a structured \emph{UV-feature residual} parameterization and to explicitly optimize the conditioning of this inversion. Concretely, we (i) restrict edit updates to a low-dimensional edit code $v$ that is decoded by a \emph{pretrained residual} UV decoder $g_\psi$ (thus constraining edits to the range of $g_\psi$), (ii) impose masked editing constraints that fit the edited region while preserving the non-edited region, and (iii) design / reweight these constraints during inversion using a spectral objective derived from a local linearization of the full decoding-and-rendering pipeline.

\begin{figure*}[t]
    \centering
    \includegraphics[width=0.95\linewidth]{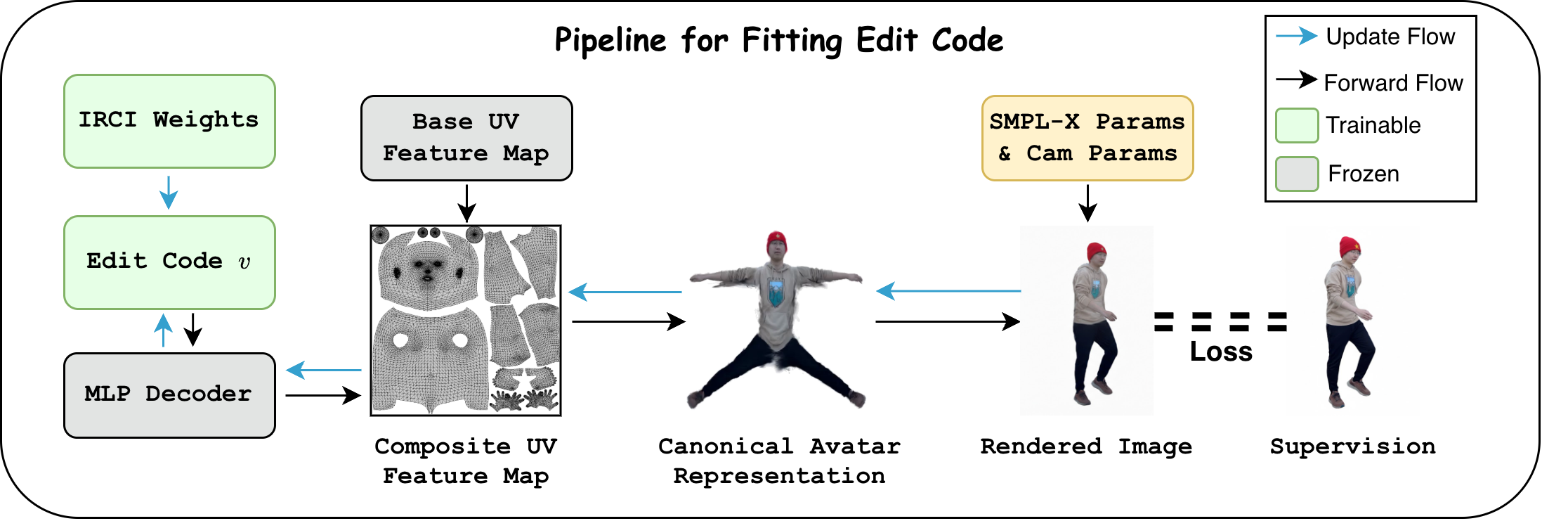}
    \caption{Overview of our pipeline.}
    \label{fig:overview}
\vspace{-0.2in}
\end{figure*}

\subsection{Differentiable Avatar Rendering Pipeline}
\label{sec:notation}

We assume a differentiable, animatable rendering pipeline
\begin{equation}
    y_t = f(v,\theta_t)\in\mathbb{R}^m,
\end{equation}
where $v\in\mathbb{R}^{r}$ is a global \emph{editing code} shared across frames,
$\theta_t$ denotes the frame-specific driving state (pose parameters, cameras, and optionally illumination),
and $y_t$ denotes rendered observations (RGB and optionally depth/normal).

\textbf{Residual UV parameterization.}
A base reconstruction produces a canonical UV feature map $U_{\text{base}}$.
We model editing as adding a decoded residual UV feature map:
\begin{equation}
    U(v) \;=\; U_{\text{base}} + g_\psi(v),
    \qquad g_\psi(\mathbf{0})=\mathbf{0}.
    \label{eq:residual_uv}
\end{equation}
Here $g_\psi$ is a \emph{pretrained and frozen} MLP decoder that maps a low-dimensional edit code
to a UV-aligned residual feature map. The constraint $g_\psi(\mathbf{0})=\mathbf{0}$
ensures that $v=\mathbf{0}$ recovers the base reconstruction exactly.

The full rendering pipeline is then
\begin{equation}
    U = U(v),\quad
    \mathcal{G}_{t}=\mathcal{A}(U,\theta_t),\quad
    y_t=\mathcal{R}(\mathcal{G}_t),
\end{equation}
i.e., $f := \mathcal{R}\circ \mathcal{A}\circ U(\cdot)$.
$\mathcal{A}$ includes canonical-to-posed deformation and Gaussian attribute prediction,
and $\mathcal{R}$ is a differentiable Gaussian splatting renderer.

We define the base rendering as
\begin{equation}
    y_t^{\text{base}} := f(\mathbf{0},\theta_t)
    \;=\; \mathcal{R}\!\left(\mathcal{A}(U_{\text{base}},\theta_t)\right).
\end{equation}

\subsection{Constrained Editing Setup}

We consider a candidate pool of frames $C\subseteq \{1,\cdots,T\}$.
For each $t\in C$, edited supervision consists of an edited target $y^{\text{edit}}_t$ and a diagonal edit mask $M_t\in\mathbb{R}^{m\times m}$ selecting the edited region. We denote $\overline{M}_t:=I-M_t$.

\textbf{Edit code.}
Instead of a linear latent subspace $\bs{z}_{\text{base}}+Pv$, we directly optimize
a low-dimensional edit code $v\in\mathbb{R}^{r}$ that is decoded into a residual UV map
by the frozen decoder $g_\psi$ in Eq.~\eqref{eq:residual_uv}.
This implicitly restricts edits to the structured manifold defined by $g_\psi$, reducing identity leakage.

For each candidate frame $t\in C$, define $\tilde{y}_t := f(v,\theta_t)$ and the per-frame loss
\begin{equation}
    \ell_t(v)
    :=
    D(M_t\tilde{y}_t,\;M_ty_t^{\text{edit}})
    +\lambda_{id}D(\overline{M}_t\tilde{y}_t,\;\overline{M}_ty_t^{\text{base}}),
    \label{eq:per-frame loss}
\end{equation}
where the distance function $D$ is a weighted sum of pixel-wise L1 RGB loss and perceptual LPIPS, and the first term fits the edited region and the second preserves non-edited regions.

\subsection{Inversion Conditioning via Edit-Code Info-Matrix}

Even with a low-dimensional edit code, sparse edited constraints may yield unstable inversions:
some edit directions remain weakly observed, leading to pose-dependent artifacts.
We diagnose and control this via a local linearization around $v=\mathbf{0}$ (the base).

We locally linearize the renderer:
\begin{equation}
    f(v,\theta_t)\approx f(\mathbf{0},\theta_t)+J_tv,
    \label{Eq:linear approx}
\end{equation}
where $J_t:=\left.\frac{\partial f(v,\theta_t)}{\partial v}\right|_{v=\mathbf{0}}\in\mathbb{R}^{m\times r}$.
The masked edit-code Jacobian is $A_t:=M_tJ_t\in\mathbb{R}^{m\times r}$.
Under a Gauss-Newton approximation, the local curvature in edit-code space is
\begin{equation}
    H_t:=A_t^\top A_t\in\mathbb{R}^{r\times r},\qquad H_t\succeq 0.
\end{equation}
\vspace{-2.5em}

Given nonnegative per-frame constraint weights $w=\{w_t\}_{t\in C}$, define the edit-code precision matrix
\begin{equation}
    S(w):=\Lambda_0 + \sum_{t\in C} w_t H_t,\quad \Lambda_0\succ 0.
\vspace{-1em}
\end{equation}
Since $r$ is small, conditioning analysis in $S(w)\in\mathbb{R}^{r\times r}$ is efficient.

\subsection{Information-Regularized Constrained Inversion}
\label{sec:ICER_objective}
We integrate inversion conditioning into edited reconstruction by jointly optimizing the edit code $v$ and the constraint weights $w$. Specifically, we minimize
\begin{align}
\min_{v,w}&\;\;
\mathcal{L}(v,w)
:=
\sum_{t\in C} w_t \ell_t(v) + v^T \Lambda_0 v  \nonumber
\\&-\lambda_{\text{cond}} \log\det\!\Big(\Lambda_0 + \sum_{t\in C} w_t H_t\Big),
\ \text{s.t. } w \in \Delta_K,
\label{eq:ICER}
\end{align}
where $\Delta_K:=\{w_t\ge0,\;\sum_{t\in C} w_t=K\}$ encodes an effective
supervision budget over candidate constraints and $v^T\Lambda_0v$ is the prior regularization term for edit vector $v$. To make \emph{keyframes} emerge from optimization rather than being heuristically pre-selected, we parameterize weights using logits $\{a_t\}$ and a temperature $\tau$: \(w_t=K\cdot \text{Softmax}(a_t/\tau)\).


Directly optimizing \eqref{eq:ICER} is nonconvex due to the renderer. We adopt an alternating scheme that is stable in practice:

\textbf{(1) Update the edit code $v$ (constrained inversion).}
With weights $w$ fixed, we minimize
\begin{equation}
\vspace{-0.5em}
    \min_{v} \;\; \sum_{t} w_t \ell_t(v) + v^T \Lambda_0 v.
\vspace{-0.5em}
    \label{eq:v_update}
\end{equation}
We optimize $v$ via conditioning-aware update,
\begin{equation}
\vspace{-0.5em}
    v \leftarrow v - \eta \, S(w)^{-1} \nabla_v \Big(\sum_t w_t \ell_t(v) + v^T \Lambda_0 v\Big),
\vspace{-0.5em}
    \label{eq:preconditioned_update}
\end{equation}
where the calculation of $S(W)^{-1}$ is efficient since $S(w)\in \mathbb{R}^{r\times r}$ and $r\ll m$.

\textbf{(2) Update the weights $w$ (inversion-time constraint design).}
With $v$ fixed, we update logits $\{a_t\}$ (hence $w$) to trade off fitting pressure and conditioning reward.
A key advantage of $\log\det S(w)$ is its closed-form gradient:
\vspace{-0.5em}
\begin{equation}
    \frac{\partial}{\partial w_t}\log\det S(w)
    =
    \mathrm{Tr}\!\left(S(w)^{-1} H_t\right).
\vspace{-0.5em}
    \label{eq:logdet_grad}
\end{equation}
Therefore, the partial derivative of \eqref{eq:ICER} w.r.t.\ $w_t$ is
\vspace{-0.5em}
\begin{equation}
    \frac{\partial \mathcal{L}}{\partial w_t}
    =
    \ell_t(v)
    - \lambda_{\rm{cond}} \, \mathrm{Tr}\!\left(S(w)^{-1} H_t\right).
\vspace{-0.5em}
    \label{eq:dL_dw}
\end{equation}
Intuitively, $\ell_t(v)$ acts as a fitting cost: frames with larger residuals are \emph{down-weighted} under gradient descent.
In contrast, the information term provides a conditioning reward with marginal gain $\mathrm{Tr}(S(w)^{-1}H_t)$, which \emph{promotes} frames that improve conditioning. Overall, the weight update reallocates the supervision budget toward frames with lower-than-average $\ell_t(v)-\lambda_{\rm{cond}}\mathrm{Tr}(S(w)^{-1}H_t)$, i.e., frames that are both easier to fit and more informative.

Since $S(w)$ is only $r\times r$, computing $\log\det S(w)$ and $\mathrm{Tr}(S^{-1}H_t)$ is efficient. It is common for a portion of the frames not to have the edited version in the inputs. For those, we simply set $\ell_t(v)=0$.

\textbf{Efficient estimation of $H_t$.} Directly forming the Jacobian $J_t$ is infeasible, but we only require the small matrix $H_t \in \mathbb{R}^{r\times r}$, defined as $H_t := A_t^\top A_t$ with $A_t := M_t J_t$ and
$J_t = \left.\frac{\partial f(v,\theta_t)}{\partial v}\right|_{v=0}$.
We compute $H_t$ via Hessian--vector products (HVPs) of the least-squares surrogate
\begin{equation}
\phi_t(v) := \tfrac12 \big\| M_t\big(f(v,\theta_t)-f(0,\theta_t)\big)\big\|_2^2 .
\end{equation}
Since the residual $r_t(v)=M_t(f(v,\theta_t)-f(0,\theta_t))$ satisfies $r_t(0)=0$,
the Hessian at $v=0$ reduces \emph{exactly} to the Gauss--Newton (Jacobian-Gram)
matrix:
\[
\nabla^2 \phi_t(0) = J_t^\top M_t^\top M_t J_t = H_t .
\]
Thus, for any $q\in\mathbb{R}^r$, $H_t q$ can be obtained by an HVP
$\nabla^2\phi_t(0)\,q$. Since $r\ll m$, we explicitly construct $H_t$ by applying
HVPs to the basis vectors $\{e_i\}_{i=1}^r$.

\textbf{Discrete keyframes.}
Although our method optimizes continuous weights, a discrete set of keyframes can be extracted for interpretability and downstream applications (e.g., interactive refinement of the reconstruction) by selecting
\(
\mathcal{K}=\text{TopK}(\{w_t\}, K).
\)

\subsection{Learning Part-specific Residual UV Decoder from Paired Mesh/Render Edits}
\label{sec:learn subspace from paired edits}

Given a dataset of paired 3D assets $\{(\mathcal{M}_i^{\text{orig}},\mathcal{M}_i^{\text{edit}})\}_{i=1}^N$ for a particular edit type $g$ (e.g., garment or shoes), where each pair shares the same identity, we pretrain a part-specific residual UV decoder $g_{\psi_g}$ and per-pair edit codes $\{v_i\}$.
Let $U_{i}^{\text{orig}}$ denote the UV feature map reconstructed from $\mathcal{M}_i^{\text{orig}}$ (serving as $U_{\text{base}}$ for that pair), and define the edited rendering under code $v_i$ by
\begin{equation}
\vspace{-0.5em}
    \tilde{y}^{(i)}_t(v_i):=\mathcal{R}\!\left(\mathcal{A}(U_{i}^{\text{orig}} + g_{\psi_g}(v_i),\theta_{i,t})\right).
\vspace{-0.2em}
\end{equation}
We train $g_{\psi_g}$ by minimizing a multi-view version of the masked edit objective (Eq.~\ref{eq:per-frame loss}) on rendered pairs:
\begin{align}
\vspace{-0.5em}
    &\min_{\psi_g,\{v_i\}}
    \sum_{i=1}^N \sum_{t\in \mathcal{V}_i}
    \Big(
    D(M^{(g)}_{i,t}\tilde{y}^{(i)}_t(v_i),\, M^{(g)}_{i,t}y^{(i),\text{edit}}_t)
    \nonumber\\ + &\lambda_{id}D(\overline{M}^{(g)}_{i,t}\tilde{y}^{(i)}_t(v_i),\, \overline{M}^{(g)}_{i,t}y^{(i),\text{orig}}_t)
    \Big)
    + \lambda_{v}\sum_i\|v_i\|_2^2,
\vspace{-0.5em}
\end{align}
where $y^{(i),\text{orig}}_t:=\mathcal{R}(\mathcal{A}(U_{i}^{\text{orig}},\theta_{i,t}))$ is the base rendering and $M^{(g)}_{i,t}$ is the part mask for edit type $g$.
We enforce $g_{\psi_g}(\mathbf{0})=\mathbf{0}$ by adding a small penalty $\|g_{\psi_g}(\mathbf{0})\|_2^2$, encouraging exact recovery of the base avatar when $v=\mathbf{0}$.

At inference time, we freeze $g_{\psi_g}$ and optimize only the edit code $v$ (and weights $w$) using our information-regularized objective Eq.~\ref{eq:ICER}.

\section{Analysis}
\label{sec:analysis}

In this section, we provide a local analysis of the \emph{information-regularized constrained inversion} objective
in Eq.~(\ref{eq:ICER}), explaining why maximizing $\log\det S(w)$ improves edit-code accuracy under inconsistent supervision.
The key message is that, under valid local linearization and noisy / inconsistent edited targets, optimizing $\log\det S(w)$
performs a D-optimal constraint design \emph{in the edit-code space}, which provably shrinks the uncertainty (and hence improves
the accuracy) of the recovered edit code.

\subsection{Local Linearization Modeling}

Recall our residual UV parameterization $U(\mathbf{v})=U_{\text{base}}+g_\psi(\mathbf{v})$ with $g_\psi(\mathbf{0})=\mathbf{0}$,
and the induced differentiable renderer $\bs{y}_t=f(\mathbf{v},\theta_t)$.
The base rendering is $\bs{y}_t^{\text{base}}=f(\mathbf{0},\theta_t)$.

For each candidate frame $t\in\mathcal{C}$, define the masked edit residual
\begin{equation}
b_t \;:=\; M_t\big(y_t^{\text{edit}}-y_t^{\text{base}}\big)\in\mathbb{R}^m.
\end{equation}
Using the local linearization in Eq.~\ref{Eq:linear approx}, let
\begin{equation}
J_t:=\left.\frac{\partial f(\mathbf{v},\theta_t)}{\partial \mathbf{v}}\right|_{\mathbf{v}=\mathbf{0}}\in\mathbb{R}^{m\times r},
\end{equation}
where $A_t:=M_t J_t\in\mathbb{R}^{m\times r}$ and $H_t:=A_t^\top A_t\succeq 0$.

We assume that the masked residual is explained by an (unknown) ground-truth edit code $v_\star\in\mathbb{R}^r$
plus inconsistency noise $\varepsilon_t$ and a (bounded) linearization remainder $r_t$:
\begin{equation}
b_t \;=\; A_t v_\star + \varepsilon_t + r_t(v_\star).
\vspace{-1em}
\label{eq:local_model}
\end{equation}

Given weights $w$, we consider the latent code $\hat v(w)$ obtained by the (linearized) weighted ridge estimator
\begin{equation}
\hat v(w)
\;:=\;
\arg\min_{v\in\mathbb{R}^r}
\sum_{t\in\mathcal{C}} w_t \|A_t v - b_t\|_2^2
\;+\; v^\top \Lambda_0 v.
\label{eq:weighted_ridge}
\end{equation}

\subsection{Main Results}
\label{sec:main_results}

Our main theorem relies on the following assumptions.

\textbf{Bounded second-order remainder.}
There exists $\beta\ge 0$ such that for all $t\in\mathcal C$ and all $v\in\mathbb R^r$,
\begin{equation}
\vspace{-0.5em}
\|r_t(v)\|_2 \;\le\; \frac{\beta}{2}\,\|v\|_2^2.
\label{eq:quad_remainder}
\end{equation}

\textbf{Inconsistent supervision as heteroscedastic Gaussian noise.}
Conditioned on weights $w=\{w_t\}_{t\in\mathcal C}$, the noise terms are independent across $t$ and satisfy
\begin{equation}
\vspace{-0.5em}
\varepsilon_t \sim \mathcal N\!\big(0,\; \frac{1}{w_t}I\big),
\label{eq:noise_precision}
\end{equation}
with the convention that $w_t=0$ removes the corresponding constraint (i.e., infinite variance).
Thus, $w_t$ directly allocates a \emph{precision budget} between candidate constraints.

\textbf{Budgeted weights and bounded curvature.}
The weights satisfy $w\in\Delta_K:=\{w_t\ge 0,\; \sum_{t\in\mathcal C} w_t = K\}$.
Moreover, $\Lambda_0\succeq \lambda_0 I$ with $\lambda_0>0$, and $\|H_t\|_2\le L$ for all $t$.

Define the edit-code information matrix
\begin{equation}
S(w)
\;:=\;
\Lambda_0 + \sum_{t\in\mathcal{C}} w_t H_t.
\label{eq:scaled_information}
\end{equation}

\begin{theorem}
\label{thm:main}
Under the assumptions above, let $\hat v(w)$ denote the solution to Eq.~\ref{eq:weighted_ridge}. Then:

\vspace{0.25em}
\noindent
\textbf{(i) Posterior precision.}
If $r_t\equiv 0$ and $v_\star\sim\mathcal N(0,\Lambda_0^{-1})$, then the posterior is Gaussian:
\begin{equation}
v_\star \mid \big(\{b_t\}_{t\in\mathcal C}, w\big)
\;\sim\;
\mathcal N\!\big(\mu(w),\; S(w)^{-1}\big),
\end{equation}
so the posterior covariance is exactly $S(w)^{-1}$.

\noindent
\textbf{(ii) MSE bound with inconsistency and linearization error.}
Under the same Gaussian prior and the true model
$b_t=A_t v_\star+\varepsilon_t+r_t(v_\star)$, we have
\begin{equation}
\mathbb E\big[\|\hat v(w)-v_\star\|_2^2\big]
\;\le\;
2\,\mathrm{Tr}\!\big(S(w)^{-1}\big)
\;+\;
\frac{\beta^2 L K^2}{2\lambda_0^2}\;\mathbb E\big[\|v_\star\|_2^4\big],
\label{eq:mse_trace_bound_new}
\end{equation}
where the expectation is over $(v_\star,\{\varepsilon_t\})$.

Moreover, for $v_\star\sim\mathcal N(0,\Lambda_0^{-1})$,
\vspace{-0.5em}
\begin{equation*}
\mathbb E\big[\|v_\star\|_2^4\big]
=
\big(\mathrm{Tr}(\Lambda_0^{-1})\big)^2 + 2\,\mathrm{Tr}(\Lambda_0^{-2})
\;\le\;
\frac{r(r+2)}{\lambda_0^2}.
\vspace{-1em}
\label{eq:fourth_moment_bound}
\end{equation*}

\noindent
\textbf{(iii) Det-opt as a surrogate for Trace-opt.}
For any SPD matrix $S\in\mathbb R^{r\times r}$, AM--GM implies
\begin{equation}
\mathrm{Tr}(S^{-1}) \;\ge\; r\,\det(S)^{-1/r}.
\end{equation}
Hence, maximizing $\det(S)$ decreases a universal lower bound on the A-optimal criterion $\mathrm{Tr}(S^{-1})$.
Moreover, if the condition number $\kappa(S):=\lambda_{\max}(S)/\lambda_{\min}(S)$ is bounded by $\kappa$, then
\begin{equation}
\mathrm{Tr}(S^{-1}) \;\le\; r\,\kappa^{(r-1)/r}\,\det(S)^{-1/r}.
\end{equation}
In our setting, since $\Lambda_0\succeq \lambda_0 I$, $w\in\Delta_K$, and $\|H_t\|_2\le L$, we have
\vspace{-1em}
\begin{equation}
\kappa(S(w))\;\le\; \frac{\lambda_{\max}(\Lambda_0)+K L}{\lambda_0} \;=:\; \kappa,
\vspace{-0.5em}
\end{equation}
which is independent of $w$. Substituting into Eq.~\ref{eq:mse_trace_bound_new} yields
\begin{equation}
\mathbb E\big[\|\hat v(w)-v_\star\|_2^2\big]
\;\le\;
C_0\,\det(S(w))^{-1/r} + C_1,
\vspace{-0.5em}
\label{eq:det_bound_new}
\end{equation}
where $C_0,C_1$ are constants independent of $w$.
\end{theorem}

\textbf{Remark (identity preservation outside the mask).}
The analysis extends to Eq.~\ref{eq:per-frame loss} by augmenting each frame as
\begin{equation*}
\tilde A_t :=
\begin{bmatrix}
M_t J_t\\
\sqrt{\lambda_{\mathrm{id}}}\,\bar M_t J_t
\end{bmatrix},
\ 
\tilde b_t :=
\begin{bmatrix}
M_t (y_t^{\mathrm{edit}}-y_t^{\mathrm{base}})\\
0
\end{bmatrix},
\end{equation*}
replacing $H_t$ by $\tilde H_t=\tilde A_t^\top \tilde A_t$.
The same $\log\det$-based conditioning conclusions follow verbatim.

\section{Experiments}
\label{sec:experiments}


We evaluate \textbf{stable edited reconstruction from sparse supervision}: given a base reconstructed avatar (latent $\bs{z}_{\text{base}}$) and a pose/camera sequence $\{\theta_t\}_t^T$, the user provides only a few \emph{visually edited keyframes} with localized masks, the goal is to recover a \textbf{single edited avatar}.
Our method optimizes an edit code in a low-dimensional part-specific subspace and jointly designs per-frame constraint weights via the conditioning objective derived from the edit-subspace information matrix $S(w)$.
\vspace{-1em}
\subsection{Datasets and Visual Edit Supervision}
\label{sec:exp:data}

It is difficult to find an off-the-shelf dataset that contains dynamic human motion with both the \textit{original} and the \textit{edited} sequences. More importantly, obtaining consistent ground truth for appearance edits is notoriously hard in real captures: once an edit is applied in the physical world, the capture conditions inevitably change, which breaks the strict frame-to-frame and view-to-view correspondence required for reliable evaluation. Therefore, we adopt two complementary evaluation settings.

\textbf{Synthetic setting with pseudo-ground-truth editing.} We reconstruct 3DGS-Avatar~\cite{qian20233dgsavatar} to reconstructed the human avatar from ZJU-MoCap~\cite{peng2021neural,fang2021mirrored}, a multiview human motion dataset. We then construct a controlled edited target by applying localized material edits in the canonical avatar space and re-rendering the edited avatar for all views and frames, yielding a perfectly aligned pseudo ground truth with consistent edits.

During evaluation, we only provide a subset of the edited sequence from a single camera pose as input to the evaluated method, and test on the held-out frames/views, which serve as the testing set. This setting enables quantitative comparisons.

\textbf{In-the-wild setting.} We use a casually captured monocular sequence~\cite{jiang2022neuman} and manually edit a few keyframes via pretrained M-LLM. We then compare different methods qualitatively in terms of how well they propagate the edits. While this setting better matches our target use case, it is impossible to obtain a true ground truth for the edited appearance.

\textbf{Baselines.}
We compare against three adapted baselines: \textit{3DGS-Avatar}~\cite{qian20233dgsavatar},
\textit{DGE}~\cite{chen2024dge}, and \textit{IDOL}~\cite{zhuang2024idolinstantphotorealistic3d}.
\textit{3DGS-Avatar} is not an editor; we use it as an \emph{edit-as-reconstruction} baseline by
fine-tuning the animatable avatar on all frames while replacing supervision on edited keyframes.
\textit{DGE} is single-view and static; we initialize from the reconstructed canonical Gaussians,
update only the shared appearance using losses aggregated over all edited keyframes, and animate
the result over the full sequence. \textit{IDOL} performs canonical UV-texture editing; we
reconstruct from a single reference frame and optimize only the canonical texture to match all
edited keyframes. We select these baselines due to the lack of fully dynamic, monocular,
keyframe-edited open-source methods; together they cover reconstruction-, 3DGS-editing-, and
canonical-texture-based paradigms.






\textbf{Metrics.} Following previous works, we evaluate the model performance in the 
following metrics: PSNR and LPIPS similarity between the rendered output $\tilde{y}_t$ and the ground-truth edited image $y^{\mathrm{edit}}_t$ \emph{within} the edited region; unintended changes \emph{outside} the edited region using LPIPS on the complement region; pose/time consistency via warping error (WE). We warp $\tilde{y}_t$ to frame $t+1$ through optical flow and compute LPIPS residuals inside the edited region.

\subsection{Main Results}
\label{sec:exp:main}

\textbf{Quantitative results.}
Table~\ref{tab:main_results} summarizes results on unseen test identities under
sparse edited supervision. We report edit fidelity (masked), leakage (outside mask), and temporal stability (warping error). Our method obtains larger gains in the harder regime ($K=4$), supporting our claim that sparse edited reconstruction is fundamentally ill-conditioned without conditioning-aware constraint design. See appendix for detailed settings.

\begin{table}[t]
\centering
\small
\setlength{\tabcolsep}{4pt}
\begin{tabular}{lccccc}
\toprule
Method &
LPIPS$_e\downarrow$ &
PSNR$_e\uparrow$ &
LPIPS$_b\downarrow$ &
WE$_e\downarrow$ &
WE$_b\downarrow$ \\
\midrule
3DGSAvatar 
& 0.162 
& 23.41 
& 0.071 
& 0.048 
& 0.036 \\
DGE 
& 0.214 
& 20.87 
& 0.132 
& 0.089 
& 0.074 \\
IDOL 
& 0.187 
& 22.05 
& 0.098 
& 0.063 
& 0.051 \\
Ours 
& \textbf{0.128} 
& \textbf{25.62} 
& \textbf{0.043} 
& \textbf{0.031} 
& \textbf{0.024} \\
\bottomrule
\end{tabular}
\caption{Results on our benchmark with sparse supervision (5 uniform keyframes). We report edit fidelity (masked), leakage (outside mask), and temporal stability (warping error). Subscripts $e$ and $b$ denote edited and background regions, respectively.}
\vspace{-0.3in}
\label{tab:main_results}
\end{table}

\begin{figure}[t]
    \centering
    \includegraphics[width=0.95\linewidth]{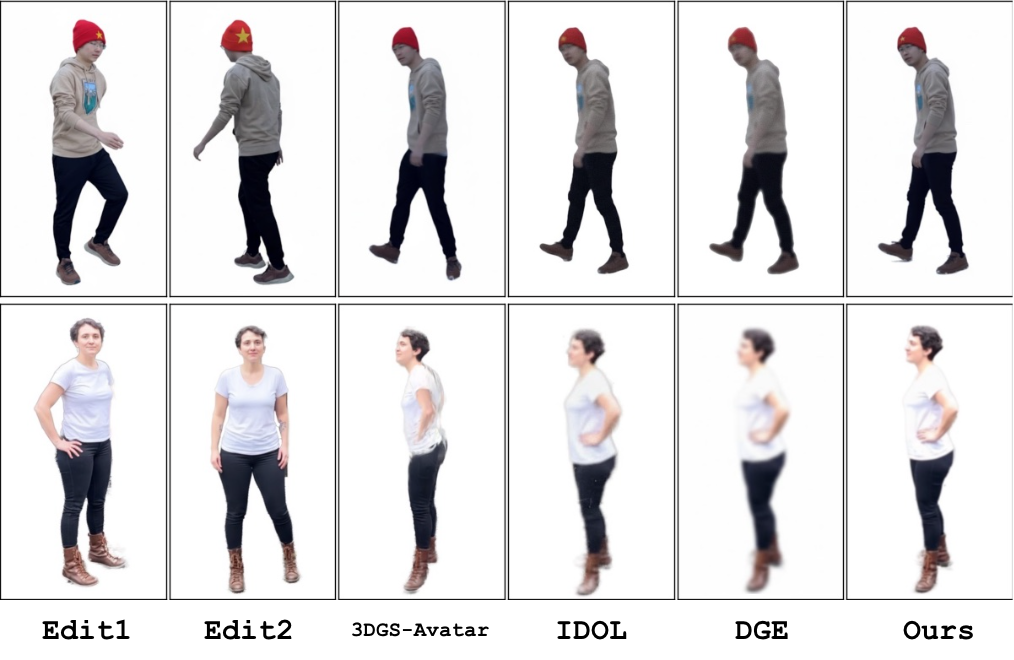}
    \caption{Edited keyframes and renderings at unseen time steps are shown. The first row contains an incorrect edit, while the second row exhibits inconsistent editing appearance, where \textbf{Edit2} introduces an arm tattoo.}
    \label{fig:qualitative}
\end{figure}

\textbf{Qualitative results.}
Figure~\ref{fig:qualitative} visualizes typical failure modes. Without conditioning-aware constraint allocation, baselines tend to overfit the few edited frames using unstable latent directions, leading to pose-dependent artifacts and temporal flicker; identity leakage is most visible on faces and unedited garments. Our method produces edits that remain consistent across pose change with minimal drift outside the edited region.

\textbf{Keyframe activation emerges from optimization.}
We visualize the learned weights $\{w_t\}$. which show that optimization
activates frames that (i) reveal informative views of the edited part and (ii)
improve the spectral conditioning of $S(w)$.
This supports that ``keyframes'' should be treated as \emph{designed
constraints} rather than a pre-processing choice.

\begin{figure}[t]
\centering
\includegraphics[width=0.95\linewidth]{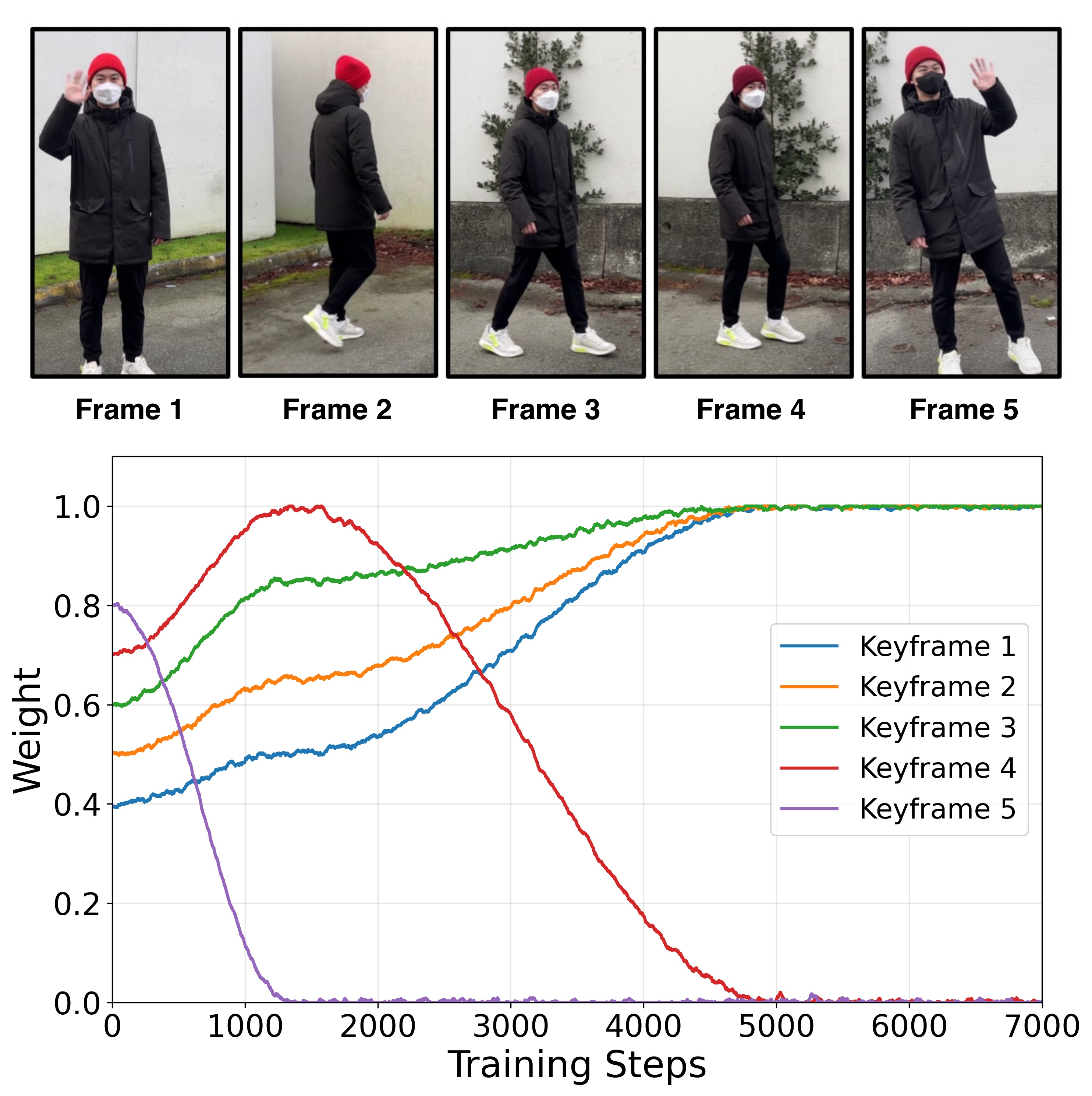}
\caption{We set the supervision budget to $K{=}3$ over five candidate keyframes. As optimization progresses, the weight mass shifts away from an obviously incorrect edit (frame~5), and subsequently from a frame with inconsistent appearance (frame~4, hat mismatch). The final selection concentrates on frames~1--3, yielding a consistent keyframe set.}
\vspace{-0.2in}
\end{figure}

\textbf{Robustness to Imperfect / Inconsistent Supervision.} In practice, edited keyframes produced by reference-driven 2D editors are often not perfectly consistent due to stochasticity, occlusions, and mask imperfections. We therefore evaluate robustness under controlled corruption of sparse supervision.

\subsection{Ablations}
\label{sec:exp:ablations}

We evaluate different components of  Eq.~\eqref{eq:ICER}.

\textbf{A1: Conditioning/weights.} Remove information regularization ($\lambda_{\text{cond}}=0$) and use uniform weights $w_t=K/|\mathcal{C}|$ to isolate the role of $\log\det S(w)$. \textbf{A2: Structured edit parameterization.} Compare our pretrained (optionally part-/edit-type-specific) residual UV decoder $g_\psi$ (i.e., $U=U_{\text{base}}+g_\psi(v)$) to (i) no restriction---directly optimizing a free UV residual $\Delta U$ with $U=U_{\text{base}}+\Delta U$, and (ii) generic parameterizations of the same code dimension, PCA linear decoder $Bv$, to probe structured control vs.\ leakage and instability. \textbf{A3: Identity term.} Set $\lambda_{\text{id}}=0$ to measure identity leakage, especially outside edited regions.

We find that the full model consistently outperforms these variants across all metrics, indicating that conditioning-aware constraint allocation, structured edit parameterization, identity preservation, and the learned prior are all critical under sparse supervision.

\begin{table}[t]
\centering
\small
\setlength{\tabcolsep}{4pt}
\begin{tabular}{lccc}
\toprule
Method &
$\text{LPIPS}_{\text{full}}\downarrow$ &
$\text{WE}_{\text{edit}}\downarrow$ &
$\log\det S(w)\uparrow$ \\
\midrule
Full & \textbf{0.082} & \textbf{0.031} & \textbf{4.21} \\
A1   & 0.095          & 0.047          & 3.05 \\
A2   & 0.111          & 0.058          & 4.07 \\
A3   & 0.107          & 0.039          & 4.13 \\
\bottomrule
\end{tabular}
\caption{Ablation study. See Sec.~\ref{sec:exp:ablations} for details.}
\vspace{-0.3in}
\label{tab:ablation}
\end{table}

\section{Conclusion, Limitations and Future Work}

We presented a conditioning-guided framework for stable edited reconstruction of animatable human avatars. Starting from a base reconstruction UV feature map $U_{\text{base}}$, we parameterize the edits with a low-dimensional code $v$ decoded by a frozen residual UV decoder $g_\psi$ (with $g_\psi(\mathbf{0})=\mathbf{0}$), and jointly optimize $v$ together with keyframe weights that improve the conditioning of the edit-code inversion system. This conditioning-aware allocation yields more reliable propagation of sparse edits across poses and time while limiting identity drift, without training a large-scale 3D editing foundation model. Our approach relies on local linearization for conditioning signals and can degrade under very large edits or noisy/inconsistent masks and targets, motivating iterative re-linearization, uncertainty-aware weighting, and improved mask estimation. Future work includes adaptive keyframe acquisition, instance-adaptive/nonlinear edit manifolds that retain Jacobian-based conditioning analysis, and tighter integration with learned priors.

The approach relies on local linearization around the base latent to derive conditioning signals; very large edits may violate this assumption, motivating iterative re-linearization or multi-stage selection strategies. Performance also depends on mask and target quality: noisy masks or view-inconsistent edited targets can degrade both inversion and keyframe design, suggesting robustness extensions such as uncertainty-aware weighting and improved mask estimation. In addition, our deformation model does not explicitly capture complex cloth dynamics or topology changes (e.g., long, highly non-rigid garments), which may require additional dynamics modeling or stronger multi-frame supervision for physically plausible geometry changes. 

Promising directions include adaptive keyframe acquisition (requesting additional edits where constraints are predicted to be most useful), nonlinear or instance-adaptive edit subspaces (replacing $P_g$ with a conditional basis while retaining local Jacobian analysis), and tighter integration with learned priors (interpreting stronger priors as shaping the effective regularization in the constrained inversion).


\nocite{langley00}

\bibliography{main}
\bibliographystyle{icml2026}

\newpage
\appendix
\onecolumn
\section{Proof for Theorem~\ref{thm:main}}
We prove items (i)--(iii) in Theorem~\ref{thm:main}. Throughout we use the local linear model
\begin{equation}
b_t = A_t v_\star + \varepsilon_t + r_t(v_\star),\qquad t\in\mathcal C,
\label{eq:model_app}
\end{equation}
and the precision-weighted noise model $\varepsilon_t\sim\mathcal N(0,\frac{1}{w_t}I)$,
independent across $t$.

\paragraph{Stacked form.}
Let $\mathcal C=\{t_1,\dots,t_n\}$ and define the stacked matrices/vectors
\begin{equation}
A_w :=
\begin{bmatrix}
\sqrt{w_{t_1}}A_{t_1}\\
\vdots\\
\sqrt{w_{t_n}}A_{t_n}
\end{bmatrix}
\in\mathbb R^{(mn)\times r},
\qquad
b_w :=
\begin{bmatrix}
\sqrt{w_{t_1}}b_{t_1}\\
\vdots\\
\sqrt{w_{t_n}}b_{t_n}
\end{bmatrix}
\in\mathbb R^{mn}.
\label{eq:stacked_def_new}
\end{equation}
Define similarly the stacked remainder and noise
\begin{equation}
r_w(v_\star):=
\begin{bmatrix}
\sqrt{w_{t_1}}r_{t_1}(v_\star)\\
\vdots\\
\sqrt{w_{t_n}}r_{t_n}(v_\star)
\end{bmatrix},
\qquad
\tilde\varepsilon :=
\begin{bmatrix}
\sqrt{w_{t_1}}\varepsilon_{t_1}\\
\vdots\\
\sqrt{w_{t_n}}\varepsilon_{t_n}
\end{bmatrix}.
\end{equation}
By construction, $\tilde\varepsilon\sim\mathcal N(0,I)$ and the stacked model is
\begin{equation}
b_w = A_w v_\star + \tilde\varepsilon + r_w(v_\star).
\label{eq:stacked_model_new}
\end{equation}
Moreover, $A_w^\top A_w=\sum_{t\in\mathcal C}w_tA_t^\top A_t=\sum_{t\in\mathcal C}w_tH_t$.

\paragraph{Closed-form solution and information matrix.}
The weighted ridge estimator in Eq.~\eqref{eq:weighted_ridge} is equivalently
\begin{equation}
\hat v(w)=\arg\min_{v\in\mathbb R^r}\ \|A_w v-b_w\|_2^2 + v^\top\Lambda_0 v.
\end{equation}
The normal equations yield
\begin{equation}
\hat v(w)=S(w)^{-1}A_w^\top b_w,
\qquad
S(w):=\Lambda_0+A_w^\top A_w=\Lambda_0+\sum_{t\in\mathcal C}w_tH_t.
\label{eq:closed_form_new}
\end{equation}

\subsection{Proof of (i): posterior precision}
Assume the linearized model $r_t\equiv 0$ and the Gaussian prior $v_\star\sim\mathcal N(0,\Lambda_0^{-1})$.
Then \eqref{eq:stacked_model_new} becomes $b_w|v_\star\sim\mathcal N(A_w v_\star,I)$.
Bayes' rule gives
\begin{align}
p(v\mid b_w)
&\propto
\exp\!\left(-\tfrac12\|b_w-A_w v\|_2^2\right)\,
\exp\!\left(-\tfrac12 v^\top\Lambda_0 v\right) \\
&\propto
\exp\!\left(
-\tfrac12\left[
v^\top(\Lambda_0+A_w^\top A_w)v -2v^\top A_w^\top b_w
\right]
\right).
\end{align}
Completing the square shows $p(v\mid b_w)$ is Gaussian with precision $S(w)$, hence posterior covariance
is $S(w)^{-1}$.

\subsection{Proof of (ii): MSE bound with noise + remainder}
Define the linear-model estimator (posterior mean/MAP under $r_w\equiv 0$)
\begin{equation}
\hat v_{\mathrm{linear}}(w):=S(w)^{-1}A_w^\top(A_w v_\star+\tilde\varepsilon),
\end{equation}
and write the actual estimator as
\begin{equation}
\hat v(w)=\hat v_{\mathrm{linear}}(w)+\Delta_r(w),
\qquad
\Delta_r(w):=S(w)^{-1}A_w^\top r_w(v_\star).
\label{eq:delta_r_new}
\end{equation}
Using $\|x+y\|^2\le 2\|x\|^2+2\|y\|^2$ and taking expectation over $(v_\star,\tilde\varepsilon)$,
\begin{equation}
\mathbb E\|\hat v(w)-v_\star\|_2^2
\le
2\mathbb E\|\hat v_{\mathrm{linear}}(w)-v_\star\|_2^2
+
2\mathbb E\|\Delta_r(w)\|_2^2.
\label{eq:mse_split_new}
\end{equation}

\paragraph{Step 1: bounding $\mathbb E\|\hat v_{\mathrm{linear}}(w)-v_\star\|^2$.}
Under $r_w\equiv 0$, $\hat v_{\mathrm{linear}}(w)$ is the posterior mean in a Gaussian linear model, hence
\begin{equation}
\mathbb E\big[(\hat v_{\mathrm{linear}}(w)-v_\star)(\hat v_{\mathrm{linear}}(w)-v_\star)^\top\big]=S(w)^{-1},
\qquad
\Rightarrow\quad
\mathbb E\|\hat v_{\mathrm{linear}}(w)-v_\star\|_2^2=\mathrm{Tr}(S(w)^{-1}).
\end{equation}

\paragraph{Step 2: bounding $\mathbb E\|\Delta_r(w)\|^2$.}
Since $\Lambda_0\succeq\lambda_0I$, we have $S(w)\succeq\Lambda_0\succeq\lambda_0I$ and thus
\begin{equation}
\|S(w)^{-1}\|_2 \le \frac{1}{\lambda_0}.
\label{eq:S_inv_bound_new}
\end{equation}
Moreover, $\|H_t\|_2\le L$ implies $\|A_t\|_2^2=\|A_t^\top A_t\|_2=\|H_t\|_2\le L$ and hence $\|A_t\|_2\le\sqrt L$.
Using the quadratic remainder bound \eqref{eq:quad_remainder},
\begin{equation}
\|r_t(v_\star)\|_2 \le \frac{\beta}{2}\|v_\star\|_2^2
\quad\Rightarrow\quad
\|r_w(v_\star)\|_2 \le \sqrt{\sum_t w_t}\cdot \frac{\beta}{2}\|v_\star\|_2^2 = \sqrt K\cdot\frac{\beta}{2}\|v_\star\|_2^2.
\end{equation}
A direct bound on $\Delta_r$ gives
\begin{align}
\|\Delta_r(w)\|_2
&=\|S(w)^{-1}A_w^\top r_w(v_\star)\|_2
\le \|S(w)^{-1}\|_2 \cdot \|A_w^\top\|_2 \cdot \|r_w(v_\star)\|_2 \nonumber\\
&\le \frac{1}{\lambda_0}\cdot \|A_w\|_2 \cdot \|r_w(v_\star)\|_2
\le \frac{1}{\lambda_0}\cdot \sqrt{\|A_w^\top A_w\|_2}\cdot \|r_w(v_\star)\|_2 \nonumber \\
&\le \frac{1}{\lambda_0}\cdot \sqrt{\Big\|\sum_t w_tH_t\Big\|_2}\cdot \|r_w(v_\star)\|_2
\le \frac{1}{\lambda_0}\cdot \sqrt{K L}\cdot \left(\sqrt K\frac{\beta}{2}\|v_\star\|_2^2\right) \nonumber\\
&= \frac{\beta K\sqrt L}{2\lambda_0}\,\|v_\star\|_2^2.
\end{align}
Squaring and taking expectation yields
\begin{equation}
\mathbb E\|\Delta_r(w)\|_2^2
\le
\frac{\beta^2 L K^2}{4\lambda_0^2}\,\mathbb E\|v_\star\|_2^4.
\label{eq:delta_r_moment_new}
\end{equation}
Combining \eqref{eq:mse_split_new} and \eqref{eq:delta_r_moment_new} proves \eqref{eq:mse_trace_bound_new}.

\paragraph{Gaussian fourth moment.}
If $v_\star\sim\mathcal N(0,\Lambda_0^{-1})$, then
$\mathbb E\|v_\star\|_2^4=(\mathrm{Tr}(\Lambda_0^{-1}))^2+2\mathrm{Tr}(\Lambda_0^{-2})$.
Using $\Lambda_0\succeq\lambda_0I$ gives $\mathrm{Tr}(\Lambda_0^{-1})\le r/\lambda_0$ and
$\mathrm{Tr}(\Lambda_0^{-2})\le r/\lambda_0^2$, hence
$\mathbb E\|v_\star\|_2^4\le r(r+2)/\lambda_0^2$.

\subsection{Trace--determinant inequality and proof of (iii)}
\begin{lemma}[Trace--determinant inequality]
Let $X\in\mathbb R^{r\times r}$ be symmetric positive definite. Then
\begin{equation}
\mathrm{Tr}(X^{-1}) \le \frac{r\,\lambda_{\max}(X)^{r-1}}{\det(X)}.
\end{equation}
\end{lemma}

Applying the lemma to $X=S(w)$ and using
$\lambda_{\max}(S(w))\le \lambda_{\max}(\Lambda_0)+\sum_t w_t\lambda_{\max}(H_t)\le \lambda_{\max}(\Lambda_0)+K L$
yields the determinant-based bound \eqref{eq:det_bound_new}.
Since the right-hand side is monotonically decreasing in $\det(S(w))$ and all other terms are independent of $w$,
any maximizer of $\log\det S(w)$ over $w\in\Delta_K$ minimizes this upper bound, proving item (iii).





\section{Dataset Details}
\subsection{Base Inversion: From a Monocular Sequence to a UV Feature Map}
\label{sec:app_uv_inversion}

Our editing pipeline operates in a structured UV-feature space and therefore requires a base UV feature map
$U_{\text{base}}$ recovered from the input monocular sequence.
We obtain $U_{\text{base}}$ by first reconstructing an animatable human avatar with a canonical 3D Gaussian representation,
and then reprojecting (``UV-aligning'') the canonical Gaussians into the UV feature-map domain.

\paragraph{Step 1: Canonical avatar reconstruction.}
Given a monocular video, we run a reconstruction method (e.g., a 3DGS-Avatar style pipeline) to obtain:
(i) a canonical set of Gaussians $\mathcal{G}^{\text{can}}=\{g_k\}_{k=1}^{N}$ (positions and appearance-related attributes),
and (ii) a pose-conditioned deformation / animation module that maps the canonical Gaussians to each frame $t$.
This step provides a stable canonical anchor for subsequent UV-space editing.

\paragraph{Step 2: Gaussian reprojection to UV space.}
For each canonical Gaussian $g_k$, we compute a UV coordinate $(u_k, v_k)$ by associating it to the parametric body
template (e.g., SMPL/SMPL-X) used in reconstruction. Concretely, we find the corresponding point on the template surface
(e.g., via nearest-point or barycentric association on the canonical template mesh) and read its UV atlas coordinate.
We then rasterize / splat Gaussian attributes into a discrete UV grid to form $U_{\text{base}}$:
\begin{equation}
U_{\text{base}}(p) \;=\;
\frac{\sum_{k} \omega_{k}(p)\, a_{k}}{\sum_{k} \omega_{k}(p) + \epsilon},
\end{equation}
where $p$ indexes a UV pixel, $a_k$ denotes the attribute/feature vector of Gaussian $g_k$ (e.g., color/opacity or learned
features), and $\omega_k(p)$ is a UV-space splatting weight (e.g., bilinear weights or a small Gaussian kernel centered at
$(u_k, v_k)$). The reprojection mapping is kept fixed during editing, so the optimization only updates UV-space features
(and/or a low-dimensional edit code) while preserving the canonical correspondence structure.

\subsection{Quantitative Benchmark Construction for Table~\ref{tab:main_results}}
\label{sec:app_benchmark}

Table~\ref{tab:main_results} reports quantitative results under \emph{sparse edited supervision} on \emph{unseen test identities}.
Since real captures do not provide perfectly aligned ``before/after'' edited ground truth, we construct a controlled benchmark
from ZJU-MoCap by combining (i) multiview capture for geometric alignment and (ii) text-prompt-based editing for realistic appearance changes.

\paragraph{Overview.}
For each identity, we start from the original ZJU-MoCap multiview sequence (images, calibrated cameras, and poses).
We select a single timestamp $t_0$ (fixed pose) as the \emph{edit acquisition instant}, and use Gemini 2.5 Flash to generate
prompt-based edits on the multiview images at $t_0$. After manual consistency filtering, we fit an edited canonical avatar to
these multiview edited images, and render the fitted edited avatar across all frames/poses to obtain temporally aligned
``ground-truth'' edited renderings for evaluation.

\paragraph{Step 1: Choose a fixed instant for editing.}
For each subject, we pick a time instant $t_0$ where the target part (e.g., coat/hat/headwear) is well visible across views.
We then collect the multiview set $\{I^{\text{orig}}_{t_0,c}\}_{c=1}^{C}$ from ZJU-MoCap at that instant.

\paragraph{Step 2: Multiview text-prompt-based editing with Gemini 2.5 Flash.}
We feed each view image $I^{\text{orig}}_{t_0,c}$ into Gemini 2.5 Flash with a scene-specific prompt (Sec.~\ref{sec:app_gemini_prompts})
to obtain candidate edited images $\{I^{\text{edit}}_{t_0,c}\}$.
Because generative editors can be stochastic and may introduce cross-view inconsistencies, we generate multiple candidates
per view and then manually select a subset that is visually consistent across viewpoints
(e.g., consistent color/material/shape of the edited part, and no unintended changes to identity/background/lighting).

\paragraph{Step 3: Fitting an edited canonical avatar to obtain ground truth.}
Using the selected multiview edited set at $t_0$, we perform a high-quality fitting step to obtain an edited canonical representation
(e.g., edited canonical Gaussians or an edited UV feature map).
In practice, we keep the original deformation/animation module (learned from the original sequence) fixed,
and optimize only the canonical appearance-related parameters to match $\{I^{\text{edit}}_{t_0,c}\}$ at the fixed pose $t_0$.
We then render this fitted edited avatar for all frames $t$ and evaluation views $c$, producing aligned ground-truth edited images
$\{I^{\text{gt}}_{t,c}\}$ for quantitative comparison.

\paragraph{Evaluation protocol under sparse supervision.}
To simulate sparse edited supervision, we provide only a small set of edited keyframes from a single input camera (monocular supervision).
We report:
(i) \emph{edit fidelity} inside the edited mask,
(ii) \emph{leakage} outside the mask (unintended changes),
and (iii) \emph{temporal stability} measured by warping error.
We highlight the harder supervision budget regime (e.g., $K=4$) in Table~\ref{tab:main_results}.

\subsection{Gemini 2.5 Flash Prompts}
\label{sec:app_gemini_prompts}

We list the prompts used to produce the edited supervision in our qualitative/quantitative scenarios:

\paragraph{(1) Coat $\rightarrow$ T-shirt.}
\begin{quote}
\footnotesize\ttfamily
Change the blue coat into white T-shirt, keep human identity, background and lighting unchanged.
\end{quote}

\paragraph{(2) Blue hat $\rightarrow$ red hat with a yellow star.}
\begin{quote}
\footnotesize\ttfamily
Change the blue hat into red hat with a yellow star on the front, keep human identity, background and lighting unchanged.
\end{quote}

\paragraph{(3) Headphone $\rightarrow$ red knitted hat (keep white mask unchanged).}
\begin{quote}
\footnotesize\ttfamily
Change the white headphone into red knitted hat, keep human identity, background and lighting unchanged, don't change the white mask.
\end{quote}


\end{document}